\documentclass{article}
\usepackage{spconf,amsmath,graphicx}
\pdfoutput=1
\usepackage{color}
\usepackage[table]{xcolor}
\usepackage{makecell,booktabs}
\usepackage{multirow}
\usepackage{CJKutf8}
\usepackage{arydshln}
\usepackage{colortbl}
\usepackage{enumitem}
\setlist{nolistsep}

\title{SLM: Bridge the thin gap between speech and text foundation models}

\name{
\begin{tabular}{cc}
Mingqiu Wang, Wei Han, Izhak Shafran, Zelin Wu, Chung-Cheng Chiu, Yuan Cao, Yongqiang Wang,\\
Nanxin Chen, Yu Zhang, Hagen Soltau, Paul K. Rubenstein, Lukas Zilka, Dian Yu, Zhong Meng \\
Golan Pundak, Nikhil Siddhartha, Johan Schalkwyk, Yonghui Wu
\end{tabular}
\thanks{Corresponding author: mingqiuwang@google.com}
\address{Google Deepmind}}

\begin{document}
%\ninept
%
\maketitle
\begin{abstract}

We present a joint Speech and Language Model (SLM), a multitask, multilingual, and dual-modal model that takes advantage of pretrained foundational speech and language models. SLM freezes the pretrained foundation models to maximally preserves their capabilities, and only trains a simple adapter with just 1\% (156M) of the foundation models' parameters. This adaptation not only leads SLM to achieve strong performance on conventional tasks such as automatic speech recognition (ASR) and automatic speech translation (AST), but also unlocks the novel capability of zero-shot instruction-following for more diverse tasks. Given a speech input and a text instruction, SLM is able to perform unseen generation tasks including contextual biasing ASR using real-time context, dialog generation, speech continuation, and question answering. Our approach demonstrates that the representational gap between pretrained speech and language models is narrower than one would expect, and can be bridged by a simple adaptation mechanism. As a result, SLM is not only efficient to train, but also inherits strong capabilities already present in foundation models of different modalities.

\end{abstract}

% \begin{keywords}
% \end{keywords}
%
\section{Introduction}
\label{sec:intro}
Recent advances in foundation models of text and speech have offered new opportunities to build strong speech-language models without a large amounts of paired speech-text data. Text foundation models have demonstrated impressive capabilities and performance on a wide range of language tasks \cite{gpt4-technical-report, anil2023palm}, and audio foundation models have recently advanced the state-of-the-art in speech recognition and understanding tasks \cite{zhang2023google, radford2023robust}. Developing effective approaches that unify foundation models of both modalities is a natural way of building strong speech understanding models without requiring a large amount of paired speech-text data.
\begin{figure}[h]
\centering
\includegraphics[width=0.5\textwidth]{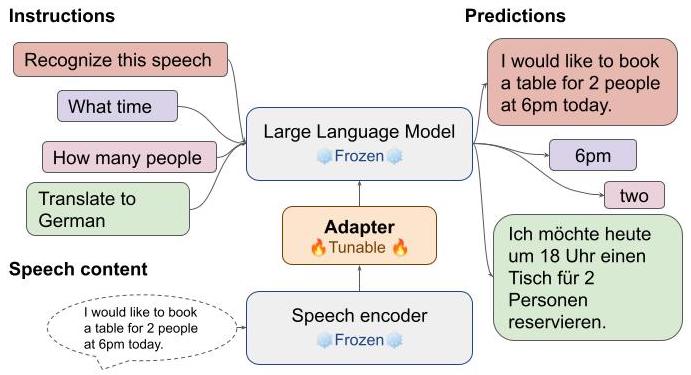}
\caption{SLM consists of a frozen pretrained speech model, a frozen pretrained LLM and an adapter to bridge from speech to textual embeddings. Therefore, SLM extends LLM's instruction-following capabilities beyond text to speech inputs and successfully performs multiple 0-shot tasks.}
\label{fig:model}
    \vspace*{-0.15in}
\end{figure}

In previous work, a joint Speech Language Model (SLM)~\cite{wang2023speech} was introduced using an adapter-based approach~\cite{houlsby_etal-2019} to unify pretrained speech and text models for an end-to-end English dialog understanding task, namely, MultiWoz~\cite{dstc11}. In this work, we refine the proposed SLM using multilingual speech and language foundation models to unlock new multitask and 0-shot capabilities. In contrast to the previous version of SLM, in this work the two foundation models are kept frozen to safeguard their inherent capabilities and an adapter is trained to bridge the two modalities. The adapter takes the output of the speech encoder, applies a uniform subsampling approach to reduce the sequence length, and learns to map the audio representation into the textual representation space that can be interpreted by the frozen LLM. 

The key contributions of this work are:

\begin{itemize}
    \item A lightweight and efficient approach to glue frozen speech and text foundation models with a simple adapter, maximally preserving the native capabilities in the pretrained foundation models.
    \item A robust and generalizable model that achieves strong performance on a variety of speech tasks including ASR, AST and speech biasing.
    \item The proposed system demonstrates novel cross-modality zero-shot instruction-following capabilities, with speech as inputs and text as instructions.
\end{itemize}

We describe our approach and model in Section~\ref{sec:adapter}, the training data and tasks in Section~\ref{sec:data_tasks}, experiment setup in Section~\ref{sec:expts}, illustrate several zero-shot capabilities in Section~\ref{sec:instruction}, and report quantitative results on ASR, AST and biasing tasks in Section~\ref{sec:results}. The implication of our results are discussed in Section~\ref{sec:discussion} and summarized in Section~\ref{sec:conclusions}.

% \pr{Note: could be interesting to compare this at a high level to flamingo. Instead of cross attention, you have adapter + injection into embeddings layer. I guess this is much simpler, which is an advantage.}
% \mq{we can mention that we will compare to flamingo style or other style adapter in future work, but since we are not allowed to publish flamingo, we won't include any results in this writting}

\section{Related Work}

To place this work in the context of the existing literature, we review a few representative models in this realm.
% https://arxiv.org/pdf/2305.11000.pdf
In SpeechGPT~\cite{zhang2023speechgpt}, the speech input is converted into HuBERT~\cite{hsu2021hubert} units which are then treated similar to the text tokens as input to the LLM. The entire model is then fine-tuned on different speech tasks. The capabilities of the model are illustrated using anecdotal examples, but the quantitative effectiveness of their method is unclear since no metrics on benchmark tasks are reported.
The approach of AudioPaLM and AudioLM ~\cite{rubenstein2023audiopalm, borsos2023audiolm} focus on pretraining a multimodal (audio and text) foundation model using an extended vocabulary with audio and text tokens to allow audio generation. In contrast, our work focuses on utilizing two frozen pretrained foundation models.
% https://arxiv.org/pdf/2305.11834.pdf
Pengi~\cite{deshmukh2023pengi} feeds audio into a frozen language model by using the output of a speech encoder, with the encodings being treated as the prefix to the standard text prompt. The focus of Pengi is on acoustic classification of sounds, emotions and music. 
% https://arxiv.org/pdf/2305.05665.pdf
ImageBind~\cite{girdhar2023imagebind} attempts to learn a joint embedding across six modalities including images, text, audio, depth and thermal data. The model expects relevant image data in the input and cannot readily learn cross-modal capabilities, for example, from audio-text paired data. 
% https://arxiv.org/pdf/2305.10790.pdf
Listen, Think, and Understand (LTU)~\cite{gong2023listen} leverages LLM's reasoning capabilities to improve acoustic scene understanding, which was trained on a large audio QA dataset. Instead of acoustic scene understanding, our work focuses on spoken language tasks such as ASR, AST and other zero-shot language tasks.
In AudioToken~\cite{yariv2023audiotoken}, an adapter is used to concatenate acoustic embeddings to text embeddings. Similar to our work, they trained only the adapter while keeping the acoustic encoder and the text-to-image diffusion model frozen. However, the focus of their work is image generation, while this work aims to improve spoken language tasks.

\section{Model: The Adapter Sandwich}
\label{sec:adapter}
\begin{figure}[h]
\centering
\includegraphics[width=0.5\textwidth]{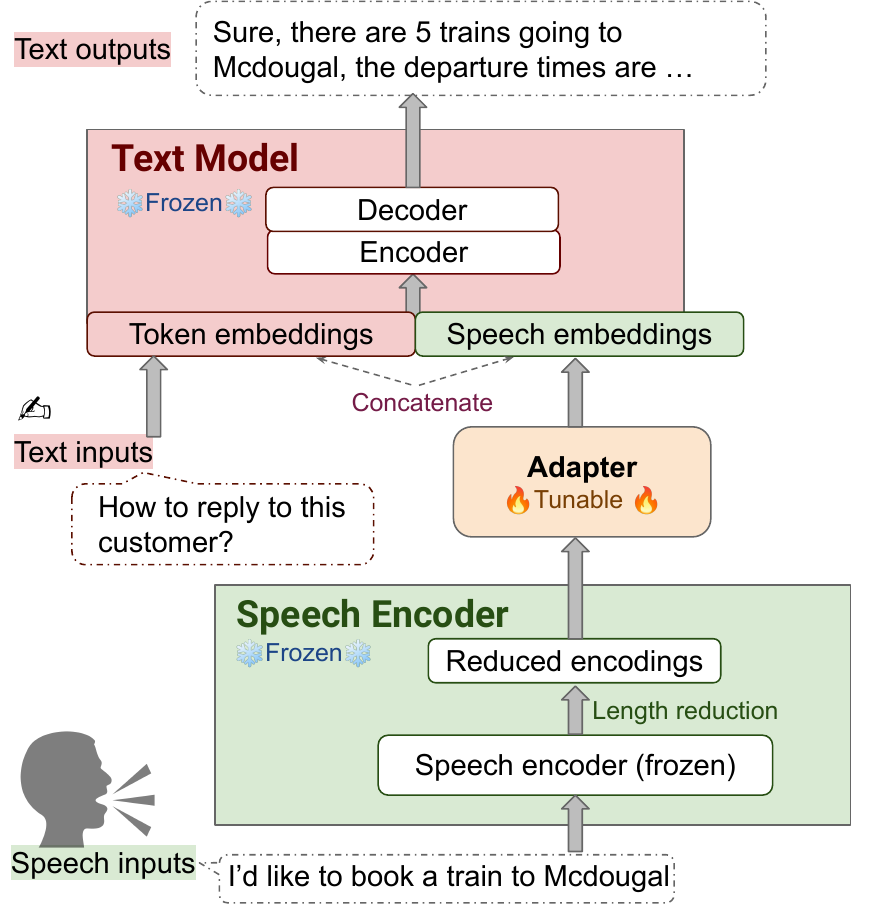}
\caption{\textbf{Model architecture of SLM.} The encodings from the output of the speech encoder is downsampled and adapted to the input textual embedding representation of the frozen LLM.}
\label{fig:model_architecture}
    \vspace*{-.15in}
\end{figure}

SLM glues a pretrained speech encoder with a pretrained LLM using a simple adapter where the adapter is sandwiched between the two frozen models, as illustrated in the Figure~\ref{fig:model_architecture}. We use SLM to refer to the combination of a pretrained LLM, a pretrained speech encoder, and the adapter. 

SLM supports two input modalities with speech and text inputs. The speech input $S_{1:U}$ of length $U$ is fed into speech encoder which generates speech embedding $S_{1:U}^{D}$ with dimension $D$. The speech encoder is taken from a pretrained encoder-decoder ASR model whose decoder is discarded. The embeddings $S_{1:U}^{D}$ are down-sampled to $S_{1:U'}^{D}$ by about a factor of 4x, as described further in Section \ref{sec:reduction}. This reduction allows longer speech inputs. The down-sampled speech embedding sequence $S_{1:U'}^{D}$ is then fed into an adapter, which in our case is simply a few transformer layers, as few as 2 layers in our experiments.

The text input $X_{1:T}$ of length $T$ is embedded by the embedding layer of the LLM. The text embedding sequence $X_{1:T}^{E}$ is then concatenated along the time dimension with the output sequence from the adapter $S_{1:U'}^{E}$ to get $X_{1:T}^{E} || S_{1:U'}^{E}$ (i.e. {\em ``\{text instruction\} \{audio\}}''). Note that the adapter output dimension $E$ as the text embedding. The concatenated embedding sequence $X_{1:T}^{E} || S_{1:U'}^{E}$ is fed into the rest of the LLM transformer stack.

We use the next-token-prediction loss as the training objective for the adapter, using a mixture of tasks (see Section \ref{sec:data_tasks}). The target sequence can be either speech transcripts, translation sentences, or any open ended generation targets. Intuitively, SLM adapter is trained to implicitly map the reduced speech embedding $S_{1:U'}^{D}$ into the same representation space as text embedding, so that the adapted speech embedding $S_{1:U'}^{E}$ can be ``understood'' by the {\em frozen} LLM. 

By supporting both speech and text inputs in the manner described above, we hypothesize that the text input serves as an effective prompt for speech inputs, allowing the model to follow instructions. In Section~\ref{sec:instruction} we demonstrate several examples tasks, lending credibility to this hypothesis.

\subsection{Speech sequence length reduction}
\label{sec:reduction}

We reduce the speech encodings sequence to similiar length as the corresponding text word-piece sequences. This is important for improving both training and inference efficiency and allows the model to handle long speech inputs. A uniform reduction is applied, where the output sequences are reduced at a fixed rate. In our experiments, we discard 3/4 of the frames randomly and thus reduce speech encoding sequence to only 1/4 of its original length.

\section{Training Data Mixtures}
% \zelin{Please reference the papers regarding the origin of the data-sets or first appeared.}

\label{sec:data_tasks}
SLM was trained using a mixture of supervised learning tasks where the inputs include speech signals, text instructions; and the output is the text string in different tasks such as ASR transcripts and speech translation sentences.
\begin{enumerate}
\itemsep 0in
    \item Speech Recognition: The fixed instruction for this task is {\em ``Recognize this speech in \{lang\}''}, where we replace {\em\{lang\}} with the actual language name for the input speech. We used multilingual YouTube corpus \cite{zhang2023google} for this task which contains 75 languages harvested from YouTube and amounts to 90k hours.
    \item Speech Translation: The model takes a speech input and generates its corresponding translation for a specified target language. The instruction for this task is {\em ``Translate this speech from \{src\_lang\} to \{tgt\_lang\}''}, where we replace {\em\{src\_lang\}} with the actual language name for the input speech, and {\em\{tgt\_lang\}} with the target language name to be translated into. We use CoVoST2 corpus \cite{wang2020covost} for this task, which is a speech to text translation dataset covering translations from 21 languages into English and from English into 15 languages, totaling about 2.9k hours of audio.
    \item Speech Instruction Tuning: The model takes a speech input and a text input as instruction, and predicts an appropriate answer following this instruction. Different from previous tasks using a fixed instruction, this task has varied instructions for different data samples such as dialog generation, named entity recognition and question answering. The task is to train the model to adeptly follow diverse instructions, avoiding over-fitting to any fixed instructions above, which is critical to the success of downstream 0-shot instruction following tasks. We used Alpaca dataset \cite{alpaca}, in which data samples contain {\em \{instruction, input, output\}}, where {\em instruction} describes the task the model should perform, {\em input} is an input context for the task, and {\em output} is the answer to the instruction. The speech input was generated using a TTS system described in ~\cite{JiaTTS2021, shenTTS2021}.
    % Note that only around 40\% data have an input. So we dropped all data samples without input, and the final speech-Alpaca dataset contains around 20K data points.
\end{enumerate}

\section{Experiments}
\label{sec:expts}

\subsection{Pretrained Foundation Models}

We used the Universal Speech Model (USM)~\cite{zhang2023google} as our speech foundation model. USM encoder of 2B parameters was first pretrained with BEST-RQ \cite{chiu2022self} and subsequently finetuned with a LAS decoder of 128M parameters on the 75-language YouTube corpus as introduced in \cite{zhang2023google}. 
%\todo{TODO: weihan, model size, 2B, model types: LAS for uniform embedding, language supported: 100+?, vocab size}

We used T5 family~\cite{raffel2020exploring} as the text foundation model, which has the encoder-decoder architecture. Specifically, we adopted mT0-MT XXL checkpoint with 13B model size that was readily available~\cite{muennighoff2022crosslingual}. This model was trained on mC4 corpus~\cite{xue2020mt5} which covers 101 languages with multilingual instruction tuning capability.

\subsection{Training}
\label{sec:train}
The adapter is a transformer stack with $L$ layers, where $L$ is a hyper-parameter. By default we use two layers of transformer as the adapter ($L=2$) unless stated otherwise. We adopt the same transformer implementation as used in mT0-MT XXL. The transformer layer size is also the same as in mT0-MT XXL, where the number of heads is 64, the head dimension is 64, the embedding dimension is 4096, and the projection layer dimension is 10240. This total parameter count of the adapter is 156M. The adapter parameters are learned from scratch.

We used a data mixture of combining all tasks described in the Section \ref{sec:data_tasks}, where all tasks have the unified input-output format:

\begin{itemize}
\itemsep 0in
    \item {\em Text prompt input}, instruction about the task to perform,
    \item {\em Speech input}, content as audio, and
    \item {\em Text response output}, target responses.
\end{itemize}

The mixing ratio is proportional to the number of data samples in each task. We used 250k multilingual sentence piece vocabulary. The training objective is the cross-entropy loss for next token prediction in a sequence, which is the standard language modeling objective. To preserve the capabilities of existing speech and text models, we froze both of the speech and text foundation models during training and only trained the adapter, as mentioned before. 

\subsection{Evaluation}
\label{sec:eval}

We first evaluated our model on conventional speech recognition and translation tasks.  For 0-shot instruction following, we demonstrate quantitative results on a contextual biasing ASR, where we instructed the model with real-time context. Furthermore, we also show empirical studies on 0-shot open-ended question answering tasks. 

\begin{enumerate}
\itemsep 0in
    \item \textit{Speech Recognition}: We evaluated on the test set of SpeechStew ASR~\cite{chan2021speechstew}, VoxPopuli ASR~\cite{wang2021voxpopuli}, and FLEURS ASR~\cite{conneau2023fleurs}. The performance was computed in terms of word error rate (WER) using the JiWER implementation~\cite{morris2004and}.
    % We normalise the text by ignoring capitalisation and punctuation before computing the WER. We also applied Whisper normalization to English recognition results, in order to have a fair comparison with the baseline models. 
    \item \textit{Speech Translation}: We evaluated on CoVoST2 AST task~\cite{wang2020covost} and report BLEU scores on X-to-En test sets using the SacreBLEU and corpusBLEU implementations~\cite{papineni2002bleu, post2018call}.
% We do not perform any normalization to the text before computing BLEU.

    \item \textit{Speech Recognition with Contextual Biasing}: This task evaluates the model's ability on recognizing speech using runtime context (i.e., named entities). We provide the model with real-time retrieved entities in the text prompts. We report WERs on the multi-context TTS corpora in~\cite{munkhdalai2023nam+}, where W\_PREFIX and WO\_PREFIX evaluate the in-domain performance: each utterance is assigned a correct bias entity + distractor entities; ANTI evaluates the out-of-domain performance: each utterance is associated with distractor entities only. The original corpora contains variants scaling from 0 to 3K bias entities assigned to each utterance. For simplicity, we combined the entities across test-set variants and constructed a single retrieval database with 4.55K bias entities in total, and scored each utterance against it. 

    \begin{itemize}
        \itemsep 0in
        \item ANTI: The transcript truths simulate the voice assistant traffic, examples include ``what’s the weather'', ``turns the lights to 100\%''. 
        \item W\_PREFIX: The transcript truths contain prefixed patterns such as ``open \$APPS'', ``call \$CONTACTS'', ``play \$SONGS''. 
        \item WO\_PREFIX: The transcript truths are entities chosen from \$APPS, \$CONTACTS, \$SONGS.  
    \end{itemize}  
    
\end{enumerate}

\section{Results}
\label{sec:results}

\subsection{Speech Recognition}
\label{sec:asr_results}

We present the results of ASR evaluation in Table \ref{tab:asr_results} on an English corpus using SpeechStew~\cite{chan2021speechstew}, as well as on multilingual corpora using Voxpopuli~\cite{wang2021voxpopuli} and FLEURS~\cite{conneau2023fleurs}. The instructions during evaluation are similar to the ones in training (e.g., {\em ``Recognize this speech in \{lang\}''}). Note that, all ASR evaluations are performed on out-of-domain tasks, since we didn't include any training data from SpeechStew, Voxpopuli, or FLEURS in the training mixture. We compare performance of our SLM model to USM baselines\cite{zhang2023google} trained on the same YouTube dataset as used in our training mixture. 

\begin{table}[h]
    \centering
    \begin{tabular}{lccc}
    \toprule
        Eval set % USM-CTC
        & USM-LAS  & SLM & SLM-FT \\ \hline
            \midrule

        \multicolumn{4}{c}{English} \\
\hdashline
        Common Voice        %& 14.3 
        & $12.6$ & $10.8$ & $7.5$ \\
        AMI (ihm)           %& 16.0
        & $16.6$ & $18.4$ & $15.4$ \\
        AMI (sdm)           %& 37.8
        & $36.3$ & $40.7$ & $36.9$\\
        Librispeech (clean) %& 5.3
        & $3.2$  &  $4.8$ & $2.6$\\
        Librispeech (other) %& 9.8
        & $5.5$  &  $7.4$ & $5.0$ \\
        Switchboard        % & 10.8
        & $10.6$ & $12.7$ & $10.3$ \\
        Tedium              %& 4.1
        &  $2.9$ &  $3.4$ & $2.9$ \\
        Wall Street Journal %& 6.5
        &  $4.8$ &  $4.4$ & $3.0$ \\
            \midrule
        \multicolumn{4}{c}{Multilingual} \\
            \hdashline
        Voxpopuli  & 13.1 &$14.0$ & $13.0$ \\
        FLEURS &13.3 & $13.8$ & $12.4$ \\
\bottomrule
    \end{tabular}
    \caption{\textbf{Speech recognition results.} For most languages, we computed Word Error Rate (WER \%) after removing capitalization, punctuation and text normalization. For Chinese, Japanese, Thai, Lao, and Burmese character error rate (CER \%) is computed similar to Whisper~\cite{radford2023robust}. Voxpoluli WER is an average of 14 languages. FLEURS WER is an average of 54 languages in Fleurs which are also present in the YouTube corpus. For SpeechStew dataset, Whisper normalization was applied on references and predictions. We also report performance of a fine-tuned SLM (SLM-FT) after training SLM text encoder on YouTube corpus.}
    \label{tab:asr_results}
\vspace{-.1in}
    
\end{table}

\begin{table}
  \label{table:top-level-results}
  %\begin{adjustwidth}{-.5in}{-.5in}  
  \centering
  \begin{tabular}{lc}
    \toprule
    \multirow{1}{*}{Model}     & BLEU$\uparrow$ (X-to-En)  \\
    \midrule
    \midrule
    Whisper \cite{radford2023robust} & $29.1$    \\
    mSLAM-CTC \cite{bapna2022mslam} & $25.2$    \\
    MAESTRO \cite{chen2022maestro} & $25.2$     \\
    USM-M \cite{zhang2023google} & $30.7$ \\
    Mu$^{2}$SLAM \cite{cheng2023mu} & $27.1$  \\
    AudioPaLM-2 \cite{rubenstein2023audiopalm} & $37.8$   \\
    \midrule
    SLM & $33.0$ \\
    SLM-FT & $37.4$ \\
    \bottomrule
  \end{tabular}
  %\end{adjustwidth}
    \caption{\textbf{Speech translation results on CoVoST2 test set.}}
    \label{tab:ast_results}
        \vspace*{-.15in}
\end{table}

\subsection{Speech Translation}
\label{sec:ast_results}

We report speech translation performance on CoVoST2~\cite{wang2020covost} corpus in Table \ref{tab:ast_results}, where the performance is averaged over 21 pairs of X-to-En translation. Here again, the instruction during evaluation and training are the same (e.g., {\em ``Translate this speech from \{src\_lang\} to \{tgt\_lang\}''}). In this case, the response from the model are scored without any normalization.

\subsection{Zero-shot Instruction Following}
\label{sec:instruction}

\subsubsection{Speech Recognition with Contextual Biasing}
\label{sec:biasing_results}
 
We evaluated the contextual biasing ASR as a general speech recognition task using the same instruction to that used in other ASR tasks, as shown in the 2nd column in Table \ref{tab:biasing_results}. Typically, a specific list of phrases is given for each speech utterance. We provided this list in the prompt and instructed the model to pick the most relevant phrase, i.e., {\em ``Recognize this speech in language English using potential mention - \{biasing entity\}''}. 
We used an off-the-shelf speech retriever \cite{wang2023speech} to retrieve top-1 entity mention from the speech, and replace {\em \{biasing entity\}} with it in the prompt above.

In the 0-shot instruction prompt experiment (C-ASR), we observe the SLM model gives about 46.2\% relative ($32.7 \rightarrow 17.6$) performance gain. We also demonstrate that further WER reductions can be achieved by fine-tuning the model parameters on task specific training corpora (C-ASR-FT): e.g., fine-tune adapter only: ($17.6 \rightarrow 7.8$), fine-tune T5 encoder: ($17.6 \rightarrow 5.1$).  

\begin{table}[h]
    \centering
    
    \begin{tabular}{lcccc} 
    \toprule

        Prompt type          & ASR  & C-ASR & \multicolumn{2}{c}{C-ASR-FT} \\
                  &   &  & (Adapter) & (T5-Enc) \\
            \midrule
\midrule
        ANTI       &  $10.3$& $10.4$ & $11.8$ & $11.2$ \\
        W\_PREFIX  &  $14.8$& $8.6$ & $1.7$ & $1.0$ \\
        WO\_PREFIX &  $32.7$& $17.6$ &  $7.8$ & $5.1$\\ 
            \bottomrule

    \end{tabular}
    \caption{\textbf{ASR contextual (C) biasing WERs.} \underline{ASR} corresponds to using the same prompt as training time {\em ``Recognize this speech in language English''}; \underline{C-ASR} corresponds to 0-shot instruction prompt; \underline{C-ASR-FT} corresponds to variants where adapter / T5-Enc model weights are further fine-tuned on task specific training corpora, which consists of 12K synthetic TTS examples with equal coverage on \$APPS, \$CONTACTS and \$SONGS, the carrier phrase patterns of the examples closely match the W\_PREFIX test-set but are combined with non-overlapping named entities for generalization.}
    \label{tab:biasing_results}
\vspace{-.2in}
    
\end{table}

\begin{table*}[h]
\vspace{-.1in}
\centering
\rowcolors{6}{gray!25}{}
\begin{tabular}{ll}
\toprule
\textbf{Instruction} & \textbf{Prediction} \\
    \midrule
\midrule
\multicolumn{2}{l}{\textbf{Audio Content}: I don't feel well since last night, I had headache and I took a pill of Ibuprofen. I went to bed early.}
\\
\hdashline

\hdashline

\multirow{2}{*}{Recognize this speech:} & I don't feel well since last night, I had headache and \\
& I took a pill of Ibuprofen. I went to bed early. \\
What symptoms are mentioned? & Headache. \\
What medicines are mentioned? & Ibuprofen. \\
What happened to this person? & This person is sick. \\
Translate this speech to Chinese: &  \begin{CJK*}{UTF8}{gbsn}从昨晚我就不太舒服，我头疼，我吃了片布洛芬。我很早睡了。 \end{CJK*} \\
\midrule
\rowcolor{white}
\multicolumn{2}{l}{\textbf{Audio Content}: can you open hay day please} \\
\hdashline

\hdashline

Recognize this speech: &  Can you open hi day please? \\
\rowcolor{gray!25} Recognize this speech with potential mentions - &\\ \rowcolor{gray!25}"Hey Ya!", "Happy Day", "Bad Day", "hay day" & Can you open hay day please? \\

\midrule
\rowcolor{white}
\multicolumn{2}{l}{\textbf{Audio Content}: Can I reserve a double-room for 4 nights, for 2 adults and a kid? Also we'd like to add breakfast.} \\
\hdashline

\hdashline
How would you answer this? & Sure, when will you be arriving? \\
How would you answer this if you don't & Sorry we don't have double-room available for 4 nights. Do you \\ \rowcolor{gray!25} have such a room? &   consider a single bed room? \\
\rowcolor{white}
\multirow{3}{*}{Translate this speech to French: } &
Est-ce que je peux réserver une chambre double pour 4 nuits, pour\\ & 2 adultes et un enfant ? \\ \rowcolor{white}& Nous aimerions également ajouter le petit-déjeuner. \\

\midrule
\midrule

\multicolumn{2}{l}{\textbf{Audio Content} (From NQ dataset): Give the formula for the following substance carbonic acid} \\
\hdashline

\hdashline

\rowcolor{white}

How do you answer this?  & H2CO3 / [\textcolor{teal}{Groundtruth: H2CO3 (equivalently OC(OH)2)}] \\

\midrule
\multicolumn{2}{l}{\textbf{Audio Content} (From NQ dataset): The resting stage of the cell cycle is} \\
\hdashline

\hdashline
\rowcolor{white}

How do you answer this?  & The phase where the cell does not divide. / [\textcolor{teal}{Groundtruth:}\\

& \textcolor{teal}{A phase where the cell has left the cycle and has stopped dividing}] \\

\bottomrule
\end{tabular}
\caption{\textbf{Zero-shot instruction following examples.} Given the same speech inputs, SLM can respond differently according to instructions. Particularly, we show predicted answers on audio Natural Question corpus. SLM is able to follow the instruction to answer the spoken question.}
\label{0-shot}
\vspace{-.1in}
\end{table*}

\subsubsection{Open-ended generation}

We prompted SLM with more diverse instructions, ranging from dialog generation, named entities recognition, and question answering (QA). See Table \ref{0-shot} for illustrative samples.

In particular, we tested the speech-based QA capabilities using Natural Question dataset~\cite{kwiatkowski2019natural}, where we verbalized the text questions into spoken versions using TTS, and prompted SLM with the instruction {\em How do you answer this?}

We observe that indeed SLM is capable of following the instruction and answering question. Like in standard LLMs, the freely-generated answers may be hallucinated. To investigate whether the hallucination was introduced from the adaptation process or inherited from the LM, we ran the original mT0-MT language model on the text-based NQ dataset, and found that similar hallucinations were present in mT0-MT for open-ended QA tasks.

\label{sec:open-ended}

\section{Discussion}
\label{sec:discussion}

\subsection{Adaptation depth}

To gain a deeper understanding of the required adapter depth for successfully integrating pretrained models from speech and language modalities, we varied the number of transformer layers in the adapter from $1$ to $8$. We observed notable performance improvement from 1 to 2 layers, but the performance saturated after 2 layers. This implies that the representation gap between the pretrained speech model and text model might be narrower than expected, and can be bridged by a shallow adaptation from speech encoding to the LLM embedding space. 

We also experiment different adaptation approaches, such as low-rank adaptation~\cite{hu2021lora} or using a more sophisticated Flamingo-style approach~\cite{alayrac2022flamingo} to inject the speech information into the LLM transformer stacks via cross-attention, which will be included in future work.

\subsection{Impact of pretrained LLMs}

We compared different LLM checkpoints in T5 family: mT5~\cite{xue2020mt5}, mT5-LM-adapted~\cite{vu2022overcoming}, mT0-MT~\cite{muennighoff2022crosslingual} (used in this work), T5~\cite{raffel2020exploring}, T5-flan~\cite{chung2022scaling}, and observed that the pretrained LLM plays a crucial role in both training efficiency and model quality after adaptation. We found that those LLMs pretrained with LM objective~\cite{bengio2000neural} (such as mT5-LM-adapted, mT0-MT, T5-flan) require significantly less time to train adaptation compared to LLMs solely pretrained with masked language model objective~\cite{devlin2018bert} (such as mT5 and T5). For example, with the same computational resources, adapting T5 or mT5 takes a few days to converge, while T5-Flan and mT0-mt takes a few hours.

The intrinsic capabilities of pretrained LLMs determine instructions following quality of the trained SLM. For LLMs without zero-shot instruction capabilities (T5 and mT5), the adapted SLM is not able to perform 0-shot instruction following either. When the LLMs have poor performances on certain downstream tasks (such as QA task), the adapted SLM also exhibits poor accuracy on those tasks.

This again confirms that the thin adapter layer itself only provides the transformation from speech modality to text modality, but does not store world-knowledge as in LLMs.

In this work, we only compared different encoder-decoder variants of LLM. However, the proposed adaptation approach also applies to decoder-only LLMs. In future work, we will present a more comprehensive comparison between both encoder-decoder and decoder-only LLMs.

\subsection{Train adapter only v.s finetune LLM}

In previous sections, we presented a general SLM for a wide range of speech tasks without the need of altering the weights of the original speech model or LLMs. In this section, we explore further finetuning SLM on any downstream corpus with LLMs unfrozen. This can be used to tailor to a specific downstream task to achieve optimal quality.

By finetuning SLM adapter using the in-domain contextual biasing training set, the WER decreases from $8.6\%$ to $1.7\%$ compared to the 0-shot case. By allowing LLM encoder unfreezing, the WER further decreases to $1.0\%$, see details in the Table \ref{tab:biasing_results}.

By finetuning SLM on CoVoST2 dataset with the LLM encoder unfrozen, we observed BLEU score increases from $33.0$ to $37.4$, which is on-par with the current SOTA CoVoST2 AST performance from AudioPaLM~\cite{rubenstein2023audiopalm}.

\subsection{End-to-end speech-to-X v.s. cascaded ASR+LLM}

A question that often comes up is whether the end-to-end model has an edge over a cascade pipeline where the speech is fed to an ASR system and the transcripts are send to LLMs. To answer this, we ran ablation studies on AST task, where we applied the same speech and text foundation model as in SLM: USM LAS model for ASR, and mT0-MT for text-to-text translation. Specifically, we prompted the mT0-MT model using a similar instruction {\em ``Translate this from \{src\_lang\} to \{tgt\_lang\}''} as we used in SLM training mixture.

We observed that the cascaded pipeline has significantly worse performance than the end-to-end SLM (i.e., CoVoST2 Fr-to-En BLEU degraded from $38$ to $32$). This is presumably due to ASR errors. One potential approach to improve the cascaded system is to further finetune LLMs on ASR transcripts, which will effectively improve the robustness to ASR errors, but will apparently requires further finetuning and alters the original capabilities of LLMs.

% \pr{I would be super interested in a small discussion on how the learned embeddings of the audio (after feeding through USM + downsampling + adapter) relate to the text tokens. There is an intuition that this method works because the adaptor makes the speech "look like text tokens", and it would be cool to quantify that somehow. For instance, if you just do k-nearest-neighbours against the text-token-embeddings, can you directly recover the text corresponding to the speech in the audio?}

\section{Conclusions}
\label{sec:conclusions}

We present SLM, a multitask, multilingual, and dual-modal speech-language model. SLM comprises a frozen pretrained speech encoder, a frozen pretrained LLM, and a light-weight adapter that maps the output of the speech encoder to the input of the LLM. Apart from the speech input, additional text input can be used as the prompts to specify the tasks that SLM needs to perform. 

In this work, we showcase the adaptation of output encodings from speech foundation model USM~\cite{zhang2023google} to input textual embeddings of large language model mT0-MT~\cite{muennighoff2022crosslingual}. Nevertheless, SLM can be easily applied as a plugin for any speech encoder and LLM pair. In future work, we will present a more comprehensive comparison across different speech encoders and both encoder-decoder and decoder-only LLMs. We will also compare different adaptation approaches, for example, residual~\cite{houlsby_etal-2019} adaptation or LoRA~\cite{hu2021lora}. 

\section{Acknowledgements}
We are grateful for help, discussion, and support from Nobuyuki Morioka, Heiga Zen, Yifan Ding, Ankur Bapna, Gang Li, Laurent El Shafey, James Qin, Jeffrey Zhao, Zhehuai Chen, and Yong Cheng.

\bibliographystyle{IEEEbib}
\bibliography{refs}

\begin{thebibliography}{10}

\bibitem{gpt4-technical-report}
OpenAI,
\newblock ``Gpt-4 technical report,'' 2023.

\bibitem{anil2023palm}
Rohan Anil, Andrew~M Dai, Orhan Firat, Melvin Johnson, Dmitry Lepikhin,
  Alexandre Passos, Siamak Shakeri, Emanuel Taropa, Paige Bailey, Zhifeng Chen,
  et~al.,
\newblock ``Palm 2 technical report,''
\newblock {\em arXiv}, 2023.

\bibitem{zhang2023google}
Yu~Zhang, Wei Han, James Qin, Yongqiang Wang, Ankur Bapna, Zhehuai Chen, Nanxin
  Chen, Bo~Li, Vera Axelrod, Gary Wang, et~al.,
\newblock ``Google usm: Scaling automatic speech recognition beyond 100
  languages,''
\newblock {\em arXiv}, 2023.

\bibitem{radford2023robust}
Alec Radford, Jong~Wook Kim, Tao Xu, Greg Brockman, Christine McLeavey, and
  Ilya Sutskever,
\newblock ``Robust speech recognition via large-scale weak supervision,''
\newblock in {\em Proc. ICML}, 2023.

\bibitem{wang2023speech}
Mingqiu Wang, Izhak Shafran, Hagen Soltau, Wei Han, Yuan Cao, Dian Yu, and
  Laurent~El Shafey,
\newblock ``Speech-to-text adapter and speech-to-entity retriever augmented
  llms for speech understanding,''
\newblock {\em arXiv}, 2023.

\bibitem{houlsby_etal-2019}
Neil Houlsby, Andrei Giurgiu, Stanislaw Jastrzebski, Bruna Morrone, Quentin
  de~Laroussilhe, Andrea Gesmundo, Mona Attariyan, and Sylvain Gelly,
\newblock ``Parameter-efficient transfer learning for nlp.,''
\newblock in {\em Proc. ICML}, 2019.

\bibitem{dstc11}
Hagen Soltau, Izhak Shafran, Mingqiu Wang, Abhinav Rastogi, Jeffrey Zhao,
  Ye~Jia, Wei Han, Yuan Cao, and Aramys Miranda,
\newblock ``Speech aware dialog system technology challenge (dstc11),''
\newblock {\em Proc. Interspeech}, 2023.

\bibitem{zhang2023speechgpt}
Dong Zhang, Shimin Li, Xin Zhang, Jun Zhan, Pengyu Wang, Yaqian Zhou, and
  Xipeng Qiu,
\newblock ``Speechgpt: Empowering large language models with intrinsic
  cross-modal conversational abilities,''
\newblock {\em arXiv}, 2023.

\bibitem{hsu2021hubert}
Wei-Ning Hsu, Benjamin Bolte, Yao-Hung~Hubert Tsai, Kushal Lakhotia, Ruslan
  Salakhutdinov, and Abdelrahman Mohamed,
\newblock ``Hubert: Self-supervised speech representation learning by masked
  prediction of hidden units,''
\newblock {\em IEEE/ACM Transactions on Audio, Speech, and Language
  Processing}, 2021.

\bibitem{rubenstein2023audiopalm}
Paul~K Rubenstein, Chulayuth Asawaroengchai, Duc~Dung Nguyen, Ankur Bapna,
  Zal{\'a}n Borsos, F{\'e}lix de~Chaumont Quitry, Peter Chen, Dalia~El Badawy,
  Wei Han, Eugene Kharitonov, et~al.,
\newblock ``Audiopalm: A large language model that can speak and listen,''
\newblock {\em arXiv}, 2023.

\bibitem{borsos2023audiolm}
Zal{\'a}n Borsos, Rapha{\"e}l Marinier, Damien Vincent, Eugene Kharitonov,
  Olivier Pietquin, Matt Sharifi, Dominik Roblek, Olivier Teboul, David
  Grangier, Marco Tagliasacchi, et~al.,
\newblock ``Audiolm: a language modeling approach to audio generation,''
\newblock {\em IEEE/ACM Transactions on Audio, Speech, and Language
  Processing}, 2023.

\bibitem{deshmukh2023pengi}
Soham Deshmukh, Benjamin Elizalde, Rita Singh, and Huaming Wang,
\newblock ``Pengi: An audio language model for audio tasks,''
\newblock {\em arXiv preprint arXiv:2305.11834}, 2023.

\bibitem{girdhar2023imagebind}
Rohit Girdhar, Alaaeldin El-Nouby, Zhuang Liu, Mannat Singh, Kalyan~Vasudev
  Alwala, Armand Joulin, and Ishan Misra,
\newblock ``Imagebind: One embedding space to bind them all,''
\newblock in {\em Proc. CVPR}, 2023.

\bibitem{gong2023listen}
Yuan Gong, Hongyin Luo, Alexander~H Liu, Leonid Karlinsky, and James Glass,
\newblock ``Listen, think, and understand,''
\newblock {\em arXiv}, 2023.

\bibitem{yariv2023audiotoken}
Guy Yariv, Itai Gat, Lior Wolf, Yossi Adi, and Idan Schwartz,
\newblock ``Audiotoken: Adaptation of text-conditioned diffusion models for
  audio-to-image generation,''
\newblock {\em arXiv}, 2023.

\bibitem{wang2020covost}
Changhan Wang, Anne Wu, Jiatao Gu, and Juan Pino,
\newblock ``{CoVoST 2 and Massively Multilingual Speech Translation},''
\newblock in {\em Proc. Interspeech}, 2021.

\bibitem{alpaca}
Rohan Taori, Ishaan Gulrajani, Tianyi Zhang, Yann Dubois, Xuechen Li, Carlos
  Guestrin, Percy Liang, and Tatsunori~B. Hashimoto,
\newblock ``Stanford alpaca: An instruction-following llama model,'' 2023.

\bibitem{JiaTTS2021}
Ye~Jia, Heiga Zen, Jonathan Shen, Yu~Zhang, and Yonghui Wu,
\newblock ``Png {BERT:} augmented {BERT} on phonemes and graphemes for neural
  {TTS},''
\newblock {\em CoRR}, 2021.

\bibitem{shenTTS2021}
Jonathan Shen, Ye~Jia, Mike Chrzanowski, Yu~Zhang, Isaac Elias, Heiga Zen, and
  Yonghui Wu,
\newblock ``Non-attentive tacotron: Robust and controllable neural tts
  synthesis including unsupervised duration modeling,''
\newblock {\em arXiv}, 2020.

\bibitem{chiu2022self}
Chung-Cheng Chiu, James Qin, Yu~Zhang, Jiahui Yu, and Yonghui Wu,
\newblock ``Self-supervised learning with random-projection quantizer for
  speech recognition,''
\newblock in {\em Proc. ICML}, 2022.

\bibitem{raffel2020exploring}
Colin Raffel, Noam Shazeer, Adam Roberts, Katherine Lee, Sharan Narang, Michael
  Matena, Yanqi Zhou, Wei Li, and Peter~J Liu,
\newblock ``Exploring the limits of transfer learning with a unified
  text-to-text transformer,''
\newblock {\em The Journal of Machine Learning Research}, 2020.

\bibitem{muennighoff2022crosslingual}
Niklas Muennighoff, Thomas Wang, Lintang Sutawika, Adam Roberts, Stella
  Biderman, Teven~Le Scao, M~Saiful Bari, Sheng Shen, Zheng-Xin Yong, Hailey
  Schoelkopf, et~al.,
\newblock ``Crosslingual generalization through multitask finetuning,''
\newblock {\em arXiv}, 2022.

\bibitem{xue2020mt5}
Linting Xue, Noah Constant, Adam Roberts, Mihir Kale, Rami Al-Rfou, Aditya
  Siddhant, Aditya Barua, and Colin Raffel,
\newblock ``mt5: A massively multilingual pre-trained text-to-text
  transformer,''
\newblock {\em arXiv}, 2020.

\bibitem{chan2021speechstew}
William Chan, Daniel Park, Chris Lee, Yu~Zhang, Quoc Le, and Mohammad Norouzi,
\newblock ``Speechstew: Simply mix all available speech recognition data to
  train one large neural network,''
\newblock {\em arXiv}, 2021.

\bibitem{wang2021voxpopuli}
Changhan Wang, Morgane Riviere, Ann Lee, Anne Wu, Chaitanya Talnikar, Daniel
  Haziza, Mary Williamson, Juan Pino, and Emmanuel Dupoux,
\newblock ``Voxpopuli: A large-scale multilingual speech corpus for
  representation learning, semi-supervised learning and interpretation,''
\newblock {\em arXiv}, 2021.

\bibitem{conneau2023fleurs}
Alexis Conneau, Min Ma, Simran Khanuja, Yu~Zhang, Vera Axelrod, Siddharth
  Dalmia, Jason Riesa, Clara Rivera, and Ankur Bapna,
\newblock ``Fleurs: Few-shot learning evaluation of universal representations
  of speech,''
\newblock in {\em IEEE Spoken Language Technology Workshop (SLT)}, 2022.

\bibitem{morris2004and}
Andrew~Cameron Morris, Viktoria Maier, and Phil Green,
\newblock ``From wer and ril to mer and wil: improved evaluation measures for
  connected speech recognition,''
\newblock in {\em Proc. International Conference on Spoken Language
  Processing}, 2004.

\bibitem{papineni2002bleu}
Kishore Papineni, Salim Roukos, Todd Ward, and Wei-Jing Zhu,
\newblock ``Bleu: a method for automatic evaluation of machine translation,''
\newblock in {\em Proceedings of the 40th annual meeting of the Association for
  Computational Linguistics}, 2002.

\bibitem{post2018call}
Matt Post,
\newblock ``A call for clarity in reporting bleu scores,''
\newblock {\em arXiv}, 2018.

\bibitem{munkhdalai2023nam+}
Tsendsuren Munkhdalai, Zelin Wu, Golan Pundak, Khe~Chai Sim, Jiayang Li, Pat
  Rondon, and Tara~N Sainath,
\newblock ``Nam+: Towards scalable end-to-end contextual biasing for adaptive
  asr,''
\newblock in {\em 2022 IEEE Spoken Language Technology Workshop (SLT)}, 2023.

\bibitem{bapna2022mslam}
Ankur Bapna, Colin Cherry, Yu~Zhang, Ye~Jia, Melvin Johnson, Yong Cheng, Simran
  Khanuja, Jason Riesa, and Alexis Conneau,
\newblock ``mslam: Massively multilingual joint pre-training for speech and
  text,''
\newblock {\em arXiv}, 2022.

\bibitem{chen2022maestro}
Zhehuai Chen, Yu~Zhang, Andrew Rosenberg, Bhuvana Ramabhadran, Pedro Moreno,
  Ankur Bapna, and Heiga Zen,
\newblock ``Maestro: Matched speech text representations through modality
  matching,''
\newblock {\em arXiv}, 2022.

\bibitem{cheng2023mu}
Yong Cheng, Yu~Zhang, Melvin Johnson, Wolfgang Macherey, and Ankur Bapna,
\newblock ``Mu2 slam: Multitask, multilingual speech and language models,''
\newblock in {\em Proc. ICML}, 2023.

\bibitem{kwiatkowski2019natural}
Tom Kwiatkowski, Jennimaria Palomaki, Olivia Redfield, Michael Collins, Ankur
  Parikh, Chris Alberti, Danielle Epstein, Illia Polosukhin, Jacob Devlin,
  Kenton Lee, et~al.,
\newblock ``Natural questions: a benchmark for question answering research,''
\newblock {\em Transactions of the Association for Computational Linguistics},
  2019.

\bibitem{hu2021lora}
Edward~J Hu, Yelong Shen, Phillip Wallis, Zeyuan Allen-Zhu, Yuanzhi Li, Shean
  Wang, Lu~Wang, and Weizhu Chen,
\newblock ``Lora: Low-rank adaptation of large language models,''
\newblock {\em arXiv}, 2021.

\bibitem{alayrac2022flamingo}
Jean-Baptiste Alayrac, Jeff Donahue, Pauline Luc, Antoine Miech, Iain Barr,
  Yana Hasson, Karel Lenc, Arthur Mensch, Katherine Millican, Malcolm Reynolds,
  et~al.,
\newblock ``Flamingo: a visual language model for few-shot learning,''
\newblock {\em Proc. Neurips}, 2022.

\bibitem{vu2022overcoming}
Tu~Vu, Aditya Barua, Brian Lester, Daniel Cer, Mohit Iyyer, and Noah Constant,
\newblock ``Overcoming catastrophic forgetting in zero-shot cross-lingual
  generation,''
\newblock {\em arXiv}, 2022.

\bibitem{chung2022scaling}
Hyung~Won Chung, Le~Hou, Shayne Longpre, Barret Zoph, Yi~Tay, William Fedus,
  Eric Li, Xuezhi Wang, Mostafa Dehghani, Siddhartha Brahma, et~al.,
\newblock ``Scaling instruction-finetuned language models,''
\newblock {\em arXiv}, 2022.

\bibitem{bengio2000neural}
Yoshua Bengio, R{\'e}jean Ducharme, and Pascal Vincent,
\newblock ``A neural probabilistic language model,''
\newblock {\em Proc. Neurips}, 2000.

\bibitem{devlin2018bert}
Jacob Devlin, Ming-Wei Chang, Kenton Lee, and Kristina Toutanova,
\newblock ``Bert: Pre-training of deep bidirectional transformers for language
  understanding,''
\newblock {\em arXiv}, 2018.

\end{thebibliography}

% \clearpage
% \setcounter{table}{0}
% \renewcommand{\thetable}{S\arabic{table}}
% \input{appendix}

\end{document}